\documentclass[a4paper, 10pt, conference]{ieeeconf}      

\IEEEoverridecommandlockouts                              

\usepackage{graphics} 
\usepackage{epsfig} 
\usepackage{mathptmx} 
\usepackage{times} 
\usepackage{amsmath} 
\usepackage{amssymb}  
\usepackage{cite}
\usepackage{wasysym} 

\usepackage{cite}

\usepackage[top=1.46in, bottom=0.79in,left=0.77in, right=0.52in ]{geometry}
    
\makeatletter
\def\@citex[#1]#2{\leavevmode
\let\@citea\@empty
\@cite{\@for\@citeb:=#2\do
{\@citea\def\@citea{,\penalty\@m\ }%
\edef\@citeb{\expandafter\@firstofone\@citeb\@empty}%
\if@filesw\immediate\write\@auxout{\string\citation{\@citeb}}\fi
\@ifundefined{b@\@citeb}{\hbox{\reset@font\bfseries ?}%
\G@refundefinedtrue
\@latex@warning
{Citation `\@citeb' on page \thepage \space undefined}}%
{\@cite@ofmt{\csname b@\@citeb\endcsname}}}}{#1}}
\makeatother

\newcommand\Tstrut{\rule{0pt}{2.0ex}}         
\newcommand\Bstrut{\rule[-0.9ex]{0pt}{0pt}} 
\newcommand\Mstrut{\rule[-0.0ex]{0pt}{0pt}}

\newcommand{\equO}[1]{(\ref{#1})}

\newcommand{\fig}[1]{\mbox{Fig. \ref{#1}}}
\newcommand{\tab}[1]{\mbox{Table \ref{#1}}}

\newcommand{\rundeKlammern}[1] {\left( #1 \right)}
\newcommand{\norm}[1]{\biggl\lVert#1\biggr\rVert}

\usepackage{tikz}
\usepackage{textcomp}
\newcommand\copyrighttext{%
  \footnotesize This work has been submitted to the IEEE for possible publication. Copyright may be transferred without notice, after which this version may no longer be accessible.}%
\newcommand\copyrightnotice{%
\begin{tikzpicture}[remember picture,overlay]%
\node[anchor=south,yshift=10pt] at (current page.south) {\fbox{\parbox{\dimexpr\textwidth-2cm}{\copyrighttext}}};%
\end{tikzpicture}%
\vspace{-10pt}%
}

\title{\LARGE \bf
Separable Convolutional LSTMs for Faster Video Segmentation
}

\author{Andreas Pfeuffer$^{1}$ and Klaus Dietmayer$^{1}$
\thanks{$^{1}$Andreas Pfeuffer, and Klaus Dietmayer are with the Institute of Measurement, Control, and Microtechnology, Ulm University, 89081 Ulm, Germany, firstname.lastname@uni-ulm.de}%
}

\begin{document}

\maketitle
\copyrightnotice
\thispagestyle{empty}
\pagestyle{empty}

\begin{abstract}
	Semantic Segmentation is an important module for autonomous robots such as self-driving cars. The advantage of video segmentation approaches compared to single image segmentation is that temporal image information is considered, and their performance increases due to this. Hence, single image segmentation approaches are extended by recurrent units such as convolutional LSTM (convLSTM) cells, which are placed at suitable positions in the basic network architecture.
	However, a major critique of video segmentation approaches based on recurrent neural networks is their large parameter count and their computational complexity, and so, their inference time of one video frame takes up to 66 percent longer than their basic version.
	Inspired by the success of the spatial and depthwise separable convolutional neural networks, we generalize these techniques for convLSTMs in this work, so that the number of parameters and the required FLOPs are reduced significantly. Experiments on different datasets show that the segmentation approaches using the proposed, modified convLSTM cells achieve similar or slightly worse accuracy, but are up to 15 percent faster on a GPU than the ones using the standard convLSTM cells.
	Furthermore, a new evaluation metric is introduced, which measures the amount of flickering pixels in the segmented video sequence. 
\end{abstract}


\section{Introduction}

	Autonomous robots such as self-driving cars and mobile house-robots require a good scene understanding of its environment. For instance, the drivable area of the robot's surrounding is often determined by means of semantic segmentation of the images delivered by the installed cameras. The main goal of semantic segmentation is to predict a class for each image pixel, e.g. it determines if the image pixel belongs to the road, to a vehicle or to the background. Usually, the cameras deliver a stream of images. 
 	These video sequences are often segmented by processing each video frame independently from each other by means of single image segmentation approaches such as ICNet \cite{Zhao_2017_ICNet_forRealTimeSemanticSegmentationOnHighResolutionImages}, PSPNet \cite{Zhao_2017_PyramidScenParsingNetwork}, and Deeplab \cite{Chen_2018_Deeplabv3p_EncoderDecoderWithAtrousSeparableConvolutionForSemanticImageSegmentation}. Therefore, object edges are usually flickering between two frames of the video sequence, and ghost objects or partly incorrectly segmented objects often occur only in a single frame, while they are classified correctly in the next frame. These errors can be avoided by the use of image information of previous frames. For instance, image information of the last frames can be considered by means of recurrent neural networks (RNN) to improve the segmentation accuracy of the current frame. Popular RNNs are Long-Short-Term Memories (LSTMs, \cite{Hochreiter_1997_LongShortTermMemory}) networks, and their extension convolutional LSTMs (convLSTMs, \cite{Shi_2015_ConvolutionalLSTMNetwork_AMachineLearningApproachForPrecipitationNowcasting}), since they can be easily trained and integrated in neural networks.
 	Different video segmentation approaches using convLSTMs \cite{Pfeuffer_2019_SemanticSegmentationOfVideoSequencesWithConvolutionalLSTMs, Simonyan_2015_VeryDeepConvolutionalNetworksForLargeScaleImageRecognition, Yurdakul_2017_SemanticSegmentationOfRGBDVideosWithRecurrentFullyConvolutionalNeuralNetworks} show that the performance increases due to the usage of temporal image information and the amount of flickering (ghost) objects and object edges can be reduced. 
 	However, there is no appropriate evaluation metric in the literature, which can measure these flickering image pixels. Evaluation metrics, such as pixelwise accuracy (acc.) or mean Intersection of Union (mIoU) are not suitable to detect flickering image pixels, since their number of errors are very small compared to the correctly classified pixels, so that the flickering has been only analyzed qualitatively until now. Therefore, the evaluation metric mean Flickering Pixels (mFP) is introduced in this work, which delivers a measure for flickering image pixels of a video sequence by means of the difference of neighboring frames.
 	
 	Another problem of using convLSTM cells for the video segmentation task is that they are computational expensive, and the number of model parameters of the neural network is increased enormously. 
 	\begin{figure}[tbp]
		\includegraphics[width=1.0\columnwidth]{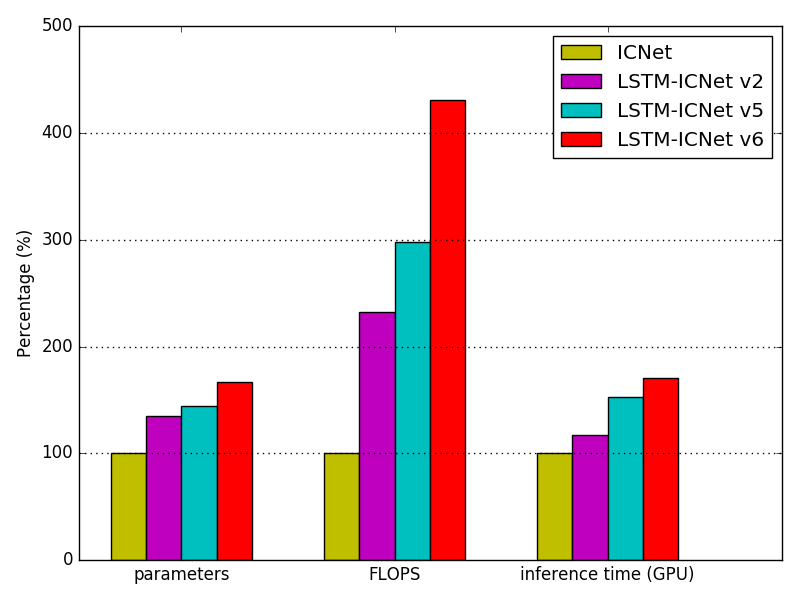}
		\caption{Comparison of the required model parameters, FLOPs and inference time on the GPU of different LSTM-ICNet versions with the original ICNet.}
		\label{fig_introduction_historgramm}
		\vspace{-5mm}
	\end{figure}
 	For example, \fig{fig_introduction_historgramm} compares the number of parameters, the required FLOPs and the corresponding inference time of different LSTM-ICNet versions proposed in \cite{Pfeuffer_2019_SemanticSegmentationOfVideoSequencesWithConvolutionalLSTMs} with the original ICNet \cite{Zhao_2017_ICNet_forRealTimeSemanticSegmentationOnHighResolutionImages}. Although the LSTM-ICNet version 2 is only extended by one convLSTM cell, the number of FLOPs increases about $133\%$, the model parameters about $18\%$ and the inference time about $35\%$ from $48ms$ to $65ms$. 
 	The other LSTM-ICNet versions containing more convLSTM cells take even longer, and their number of parameters is much greater.
 	Inspired by the popular acceleration techniques of standard convolution layers, different possibilities are introduced and discussed in this work to speed up the convLSTM cells and to reduce the large parameter count at the same time to make convLSTMs more suitable for real-time video segmentation.


\section{Related Work}


	Nowadays, recurrent neural networks (RNNs) are successfully applied in several applications, such as speech-recognition \cite{Geiger_XXXX_RobustSpeechRecognitionUsingLongShortTermMemoryRecurrentNeuralNetworksForHybridAcousticModelling, Liu_XXXX_DeepConvolutionalAndLSTMNeuralNetworksForAcousticModellingInAutomaticSpeechRecognition}, machine translation \cite{Sutsekever_2014_SequenceToSequenceLearningWithNeuralNetworks}, and hand-writing recognition \cite{Graves_2013_GeneratingSequencesWithRecurrentNeuralNetworks}. The success is based on the introduction of the Long-Short-Term Memory networks (LSTMs \cite{Hochreiter_1997_LongShortTermMemory}) in 1997, which overcomes the vanishing/exploding gradients problem of the classical RNNs. The LSTM cell consists of a memory cell to save state information. During training, the cell learns when to read the memory and when to erase or write to it.    
	However, LSTMs are computational costly and memory-excessive.
	Therefore, there are different approaches in the literature to overcome this problem. For instance, in \cite{Kuchaiev_2017_FactorizationTricksForLSTMnetworks}, two different possibilities are introduced to make LSTM cells more computationally efficient and to reduce the number of parameters at the same time. Motivated by the AlexNet \cite{Szegedy_2014_GoingDeeperWithConvolutions}, the authors partition the LSTM cell in small independent feature groups, and the output of each group is concatenated at the end to a common feature map. Furthermore, they propose a Factorized LSTM cell, in which the weight matrix $\mathbf{W}$ of size $2n \times 4n$ is approximated by the matrix product of two smaller matrices $\mathbf{W} \approx \mathbf{W}_1 \cdot \mathbf{W}_2$, where $\mathbf{W}_1$ is of size $2n \times r$ and $\mathbf{W}_2$ is $r \times 4n$, $n$ denotes the cell size and $r$ is chosen so that $\mathbf{W}$ is well approximated. In \cite{Kuchaiev_2017_FactorizationTricksForLSTMnetworks}, $r$ was set to $512$ using a cell size of $n = 8192$.
	In 2015, Shi et al. \cite{Shi_2015_ConvolutionalLSTMNetwork_AMachineLearningApproachForPrecipitationNowcasting} proposed the convolutional LSTMs (convLSTMs), which are a generalization of LSTMs for image processing tasks. Their advantage is that they are translational invariant analogously to convolution layers and the required model parameters can be significantly reduced.
	Nevertheless, convLSTMs are still time and memory expensive, and hence, different possibilities are discussed in this work how to reduce the computational costs and the number of parameters further. 
	

	A common way to accelerate standard convolution layers in neural networks is to use spatially separable convolutions, e.g. in \cite{Szegedy_2015_RethinkingtheInceptionArchitectureforComputerVision}, a \mbox{$n \times n$} convolution layer of a neural network is approximated by a \mbox{$n \times 1$} convolution layer followed by a \mbox{$1 \times n$} convolution layer, which reduces the number of parameters and FLOPs by $33\%$ in case of $n = 3$. 
	Furthermore, convolution layer can also be separated depthwise, such as in \cite{Howard_2017_MobileNets_EfficientConvolutionalNeuralNetworksForMobileVisionApplications}. In this case, each input channel is convolved independently with one filter and the amount of FLOPs and parameters is reduced enormously compared to the standard convolution. 
	In \cite{Chollet_2016_Xception_DeepLearningWithDepthwiseSeparableConvolutions, Howard_2017_MobileNets_EfficientConvolutionalNeuralNetworksForMobileVisionApplications}, a $1 \times 1 $ (pointwise) convolution is applied after the depthwise convolution, to combine the outputs of the depthwise layers. This combination of depthwise convolution and $1 \times 1$ convolution is called (depthwise) separable convolution in literature.
	These acceleration techniques of standard convolutions are generalized and applied to convLSTMs in the following sections, so that the number of parameters and the required FLOPs are reduced significantly.


	In the literature, there are only a few approaches using recurrent neural networks (RNNs) for video segmentation, since the video sequences are usually split into its individual frames, which are processed independently of each other by state-of-the-art segmentation approaches such as ICNet \cite{Zhao_2017_ICNet_forRealTimeSemanticSegmentationOnHighResolutionImages}, PSPNet \cite{Zhao_2017_PyramidScenParsingNetwork} or Deeplab \cite{Chen_2018_Deeplabv3p_EncoderDecoderWithAtrousSeparableConvolutionForSemanticImageSegmentation}. However, temporal image information is not considered in these works, but which can improve the segmentation accuracy further, as Valipour et al. show in \cite{Valipour_2017_RecurrentFullyConvolutionalNetworksForVideoSegmentation}. The authors showed on several datasets that the performance of the Fully Convolutional Network (FCN) \cite{Long_2015_FullyConvolutionalNetworksForSemanticSegmentation} can be increased if a recurrent unit is placed between the encoder and the decoder. 
	A similar approach was proposed in \cite{Yurdakul_2017_SemanticSegmentationOfRGBDVideosWithRecurrentFullyConvolutionalNeuralNetworks}, where a modified VGG19 architecture \cite{Simonyan_2015_VeryDeepConvolutionalNetworksForLargeScaleImageRecognition} was used instead of the FCN. Furthermore, different recurrent structures such as convRNN, convGRU, and convLSTM were compared in this work. It turned out that the convLSTM cells achieve the greatest accuracy.
	In \cite{Nabavi_2018_FutureSemanticSegmentationWithConvolutionalLSTM}, skip-connections containing a convLSTM cell were added to the encoder-LSTM-decoder architecture.
	In recent years, state-of-the-art approaches are rarely based on the classical encoder-decoder architecture, but use multi-branch architectures instead. Hence, the LSTM-ICNet was introduced in \cite{Pfeuffer_2019_SemanticSegmentationOfVideoSequencesWithConvolutionalLSTMs}, in which the ICNet \cite{Zhao_2017_ICNet_forRealTimeSemanticSegmentationOnHighResolutionImages} was extended by LSTM cells at suitable positions. According to \cite{Pfeuffer_2019_SemanticSegmentationOfVideoSequencesWithConvolutionalLSTMs}, the different LSTM-ICNet versions outperform the pure CNN-based approach up to $1.5$ percent, but their inference time increases up to $32ms$, which corresponds to an increase of about $66$ percent. 
	In the following work, the LSTM-ICNet versions are sped up by means of the proposed separable convLSTMs, and their inference time decreases significantly.


\section{Separable Convolutional LSTMs }

	\begin{table}[t]
		\caption{FLOPs of a standard convLSTM layer}
		\centering
		\begin{center}
			\begin{tabular}{|c|c|c|}
				\hline
				operation & total & FLOPs  \Tstrut \Bstrut \\
				\hline \hline \Tstrut
				Convolutions & $8$ & $\mathbf{8} \cdot 2 \cdot K_x \cdot K_y \cdot I \cdot  O \cdot D_x \cdot D_y$ \Mstrut\\
				Hadamard Product & $3$ & $\mathbf{3} \cdot O \cdot D_x \cdot D_y$ \Mstrut\\
				Sigma Operation & $3$ & $\mathbf{3} \cdot 5 \cdot O \cdot D_x \cdot D_y$ \Mstrut\\
				TanH Operation & $2$ & $\mathbf{2} \cdot 5 \cdot O \cdot D_x \cdot D_y$ \Mstrut\\
				Additions & $9$ & $\mathbf{9}  \cdot O \cdot D_x \cdot D_y$ \Bstrut \\
				\hline \Tstrut
				\textbf{total} & & $\left( 16 \cdot K_x \cdot K_y \cdot I + 37 \right) \cdot O \cdot D_x \cdot D_y$ \Bstrut \\
				\hline
			\end{tabular}
		\end{center}
		\label{table_FLOPS_standardConvLSTM}
		\vspace{-5mm}
	\end{table}
	
	Standard convLSTMs \cite{Shi_2015_ConvolutionalLSTMNetwork_AMachineLearningApproachForPrecipitationNowcasting} are very popular for capturing temporal information in data sequences. They use convolutional layers in their input-to-state and state-to-state transitions instead of fully connected layers, as conventional LSTMs \cite{Hochreiter_1997_LongShortTermMemory} do. The output $\mathbf{H}_t$ and the cell-states of the convLSTM layer are calculated by: 
	\begin{align}\begin{aligned}
		\mathbf{I}_t &= \sigma \rundeKlammern{\mathbf{W}_{xi} \ast \mathbf{X}_t + \mathbf{W}_{hi} \ast \mathbf{H}_{t-1} + \mathbf{b}_i}\\
		\mathbf{F}_t &= \sigma \rundeKlammern{\mathbf{W}_{xf} \ast \mathbf{X}_t + \mathbf{W}_{hf} \ast \mathbf{H}_{t-1} + \mathbf{b}_f}\\
		\mathbf{J}_t &= \tanh \rundeKlammern{\mathbf{W}_{xc} \ast \mathbf{X}_t + \mathbf{W}_{hc} \ast \mathbf{H}_{t-1} + \mathbf{b}_c}\\
		\mathbf{O}_t &= \sigma \rundeKlammern{\mathbf{W}_{xo} \ast \mathbf{X}_t + \mathbf{W}_{ho} \ast \mathbf{H}_{t-1} + \mathbf{b}_o}\\
		\mathbf{C}_t &= \mathbf{F}_t \circ \mathbf{C}_{t-1} + \mathbf{I}_t \circ \mathbf{J}_t\\
		\mathbf{H}_t &= \mathbf{O}_t \circ \tanh(\mathbf{C}_t)
		\label{equ_standardConvLSTM}
	\end{aligned}\end{align}
	where $\ast$ denotes the convolution operator and $\circ$ means the Hadamard product. The $\mathbf{W}$-terms are weight matrices, and the $\mathbf{b}$-terms are biases. $\mathbf{X}$ denotes the input of the convLSTM cell, and $\mathbf{H}$ the corresponding output, while $\mathbf{O}$ and $\mathbf{C}$ are the output state and the cell state respectively. Note, that the convLSTM cell of \equO{equ_standardConvLSTM} does not contain peephole connections like \cite{Shi_2015_ConvolutionalLSTMNetwork_AMachineLearningApproachForPrecipitationNowcasting}. However, the extension of \equO{equ_standardConvLSTM} and the remaining equations with peephole connections are straight forward, and thus, they are neglected for the sake of brevity.
	All in all,
	\begin{align}
		\left( 16 \cdot K_x \cdot K_y \cdot I + 37 \right) \cdot O \cdot D_x \cdot D_y
	\end{align}
	FLOPs are necessary using a kernel size of $K_x \times K_y$ and a feature map of size $D_x \times D_y$, if assumed that the activation functions sigma and tanh takes 5 FLOPs. The number of input channels is $I$, and the number of output channels is $O$. A detailed list of the individual components of one convLSTM cell is given in Table \ref{table_FLOPS_standardConvLSTM}. 
	
	The disadvantage of convLSTMs are their vast computational costs and their large memory consumption. For instance, LSTM-ICNet version 2, which contains only one convLSTM layer before the softmax operation, takes about $18ms$ longer on a GPU, which corresponds to an increase in inference time of about $35\%$. Therefore, three different possibilities are described on how to reduce the computational costs of LSTM-ICNet. 

\subsection{Spatially Separable Convolutional LSTMs}

	One possibility to reduce the number of FLOPs and the number of parameters is to replace a $n \times n$ convLSTM layer by a $n \times 1$ convLSTM layer followed by a $1 \times n$ convLSTM layer analogously to the Inception V3 modules \cite{Szegedy_2015_RethinkingtheInceptionArchitectureforComputerVision}. However, the convLSTM layers do not only consist of convolutions but also of other costly operations such as activation functions or elementwise multiplications (see \equO{equ_standardConvLSTM} and Table \ref{table_FLOPS_standardConvLSTM} for more details). 
	These costly operations have to be applied twice in this case. Hence, a more efficient way is to perform the spatial separation inside instead of outside of the convLSTM cell, so that the remaining operations are only executed once. In other words, each convolution $\mathbf{W} \ast \mathbf{Y}$ inside the convLSTM cell is approximated by \mbox{$\mathbf{W}^w \ast (\mathbf{W}^h \ast \mathbf{Y})$}, where $\mathbf{W}$ is a \mbox{$K_x \times K_y$} filter-kernel, and $\mathbf{W}^h$ and $\mathbf{W}^w$ are \mbox{$K_x \times 1$} and \mbox{$1 \times K_y$} filter-kernels, respectively. 
	The corresponding key equations of the spatially separable convLSTM (spatial-convLSTM) are
	\begin{align}\begin{aligned}
		\mathbf{I}_t &= \sigma \rundeKlammern{\mathbf{W}^w_{xi} \ast (\mathbf{W}^h_{xi} \ast \mathbf{X}_t) + \mathbf{W}^w_{hi} \ast (\mathbf{W}^h_{hi} \ast \mathbf{H}_{t-1}) +  \mathbf{b}_i}\\
		\mathbf{F}_t &= \sigma \rundeKlammern{\mathbf{W}_{xf}^w \ast (\mathbf{W}^h_{xf} \ast \mathbf{X}_t) + \mathbf{W}^w_{hf} \ast (\mathbf{W}^h_{hf} \ast \mathbf{H}_{t-1}) +  \mathbf{b}_f}\\
		\mathbf{J}_t &= \tanh \rundeKlammern{\mathbf{W}^w_{xc} \ast (\mathbf{W}^h_{xc} \ast \mathbf{X}_t) + \mathbf{W}^w_{hc} \ast (\mathbf{W}^h_{hc} \ast \mathbf{H}_{t-1}) + \mathbf{b}_c}\\
		\mathbf{O}_t &= \sigma \rundeKlammern{\mathbf{W}^w_{xo} \ast (\mathbf{W}^h_{xo} \ast \mathbf{X}_t) + \mathbf{W}^w_{ho} \ast (\mathbf{W}^h_{ho} \ast \mathbf{H}_{t-1}) +  \mathbf{b}_o}\\
		\mathbf{C}_t &= \mathbf{F}_t \circ \mathbf{C}_{t-1} + \mathbf{I}_t \circ \mathbf{J}_t\\
		\mathbf{H}_t &= \mathbf{O}_t \circ \tanh(\mathbf{C}_t)
		\label{equ_spatialConvLSTM}
	\end{aligned}\end{align}
	Spatial convLSTMs have the computational cost of:
	\begin{align}
		\left(32 \cdot K_x \cdot I + 37 \right) \cdot O \cdot D_x \cdot D_y,
	\end{align}
	if $K_x = K_y$. 
	By using spatial-convLSTMs instead of a standard convLSTMs, the necessary computational expenses of one cell are reduced to 
	\begin{align}
		\frac{\left(32 \cdot K_x \cdot I + 37 \right) \cdot O \cdot D_x \cdot D_y}{\left( 16 \cdot K_x \cdot K_x \cdot I + 37 \right) \cdot O \cdot D_x \cdot D_y} \approx \frac{2}{K_x}.
	\end{align}
	In case of $K_x = K_y = 3$ and $I = O = 128$, a speed-up of $66.73\%$ can be yielded in theory.

\subsection{Depthwise Convolutional LSTMs}

	The usage of depthwise convolutions instead of standard convolutions is an efficient way to reduce the computational costs, as the MobileNets  \cite{Howard_2017_MobileNets_EfficientConvolutionalNeuralNetworksForMobileVisionApplications} versions show. Hence, this concept is adapted to speed up the convLSTM layers. More concretely, all eight convolutions inside a convLSTM layer are replaced by depthwise convolutions, so that the depthwise convLSTM (depth-convLSTM) can be written as:
	\begin{align}\begin{aligned}
		\mathbf{I}_t &= \sigma \rundeKlammern{\mathbf{W}_{xi} \circledast \mathbf{X}_t + \mathbf{W}_{hi} \circledast \mathbf{H}_{t-1} + \mathbf{b}_i}\\
		\mathbf{F}_t &= \sigma \rundeKlammern{\mathbf{W}_{xf} \circledast \mathbf{X}_t + \mathbf{W}_{hf} \circledast \mathbf{H}_{t-1} + \mathbf{b}_f}\\
		\mathbf{J}_t &= \tanh \rundeKlammern{\mathbf{W}_{xc} \circledast \mathbf{X}_t + \mathbf{W}_{hc} \circledast \mathbf{H}_{t-1} + \mathbf{b}_c}\\
		\mathbf{O}_t &= \sigma \rundeKlammern{\mathbf{W}_{xo} \circledast \mathbf{X}_t + \mathbf{W}_{ho} \circledast \mathbf{H}_{t-1} + \mathbf{b}_o}\\
		\mathbf{C}_t &= \mathbf{F}_t \circ \mathbf{C}_{t-1} + \mathbf{I}_t \circ \mathbf{J}_t\\
		\mathbf{H}_t &= \mathbf{O}_t \circ \tanh(\mathbf{C}_t)
		\label{equ_depthwiseConvLSTM}
	\end{aligned}\end{align}
	where $\circledast$ denotes the depthwise convolution operator. A great advantage of depth-convLSTMs is that the number of parameters and the required FLOPs decrease enormously compared to standard convLSTMs and also to spatial-convLSTMs. For instance, the amount of FLOPs is only
	\begin{align}
		\left(16 \cdot K_x \cdot K_y + 37 \right) \cdot O \cdot D_x \cdot D_y,
	\end{align}
	which results in a theoretical speed-up of 
	\begin{align}
		\frac{\left(16 \cdot K_x \cdot K_y + 37 \right) \cdot O \cdot D_x \cdot D_y}{\left( 16 \cdot K_x \cdot K_y \cdot I + 37 \right) \cdot O \cdot D_x \cdot D_y} \approx \frac{1}{I}.
	\end{align}
	In case of $K_x = K_y = 3$ and $I = O = 128$, the computational costs amount to only $0.98\%$ of the standard convLSTM ones.

\subsection{Depthwise Separable Convolutional LSTMs}

	A disadvantage of depth-convLSTM is that cross-channel information within the convLSTM cell are not used.
	Similarly to standard depthwise convolutions, this problem can be solved by applying $1 \times 1$ convolutions after each depthwise convolution inside the convLSTM layer, so that the cross-channel information are again considered.
	More concretely, each convolution within the convLSTM cell is replaced by a (depthwise) separable convolution, and hence, for the depthwise separable convLSTM (sep-convLSTM) it holds:
	\begin{align}\begin{aligned}
		\mathbf{I}_t &= \sigma \rundeKlammern{\mathbf{W}^{1\times1}_{xi} \ast (\mathbf{W}_{xi} \circledast \mathbf{X}_t) + \mathbf{W}^{1\times1}_{hi} \ast (\mathbf{W}_{hi} \circledast \mathbf{H}_{t-1}) + \mathbf{b}_i}\\
		\mathbf{F}_t &= \sigma \rundeKlammern{\mathbf{W}_{xf}^{1\times1} \ast (\mathbf{W}_{xf} \circledast \mathbf{X}_t) + \mathbf{W}^{1\times1}_{hf} \ast (\mathbf{W}_{hf} \circledast \mathbf{H}_{t-1}) + \mathbf{b}_f}\\
		\mathbf{J}_t &= \tanh \rundeKlammern{\mathbf{W}^{1\times1}_{xc} \ast (\mathbf{W}_{xc} \circledast \mathbf{X}_t) + \mathbf{W}^{1\times1}_{hc} \ast (\mathbf{W}_{hc} \circledast \mathbf{H}_{t-1}) + \mathbf{b}_c}\\
		\mathbf{O}_t &= \sigma \rundeKlammern{\mathbf{W}^{1\times1}_{xo} \ast (\mathbf{W}_{xo} \circledast \mathbf{X}_t) + \mathbf{W}^{1\times1}_{ho} \ast (\mathbf{W}_{ho} \circledast \mathbf{H}_{t-1}) + \mathbf{b}_o}\\
		\mathbf{C}_t &= \mathbf{F}_t \circ \mathbf{C}_{t-1} + \mathbf{I}_t \circ \mathbf{J}_t\\
		\mathbf{H}_t &= \mathbf{O}_t \circ \tanh(\mathbf{C}_t),
		\label{equ_separableConvLSTM}
	\end{aligned}\end{align}
	where the $\mathbf{W}^{1\times1}$-terms are the corresponding weight matrices of the $1 \times 1$ convolution.
	In contrast to depth-convLSTM, sep-convLSTMs are computational more costly and take about
	\begin{align}
		\left(16 \cdot K_x \cdot K_y + 16 \cdot I + 37 \right) \cdot O \cdot D_x \cdot D_y,
	\end{align}
	FLOPs. Nevertheless, they are still much more efficient than standard convLSTM, and spatial-convLSTMs and only 
	\begin{align}
		\frac{\left(16 \cdot K_x \cdot K_y + 16 \cdot I + 37 \right) \cdot O \cdot D_x \cdot D_y}{\left( 16 \cdot K_x \cdot K_y \cdot I + 37 \right) \cdot O \cdot D_x \cdot D_y} \approx \frac{1}{I} + \frac{1}{K_x \cdot K_y}
	\end{align}
	of the FLOPs of one standard convLSTM cell are necessary. For example, it only takes $12.1\%$ of the computational costs of the standard convLSTM in case of $K_x = K_y = 3$ and \mbox{$I = O = 128$}.


\section{Evaluation Metric Flickering Pixels}

\begin{figure*}[t]
		\includegraphics[width=1.0\textwidth]{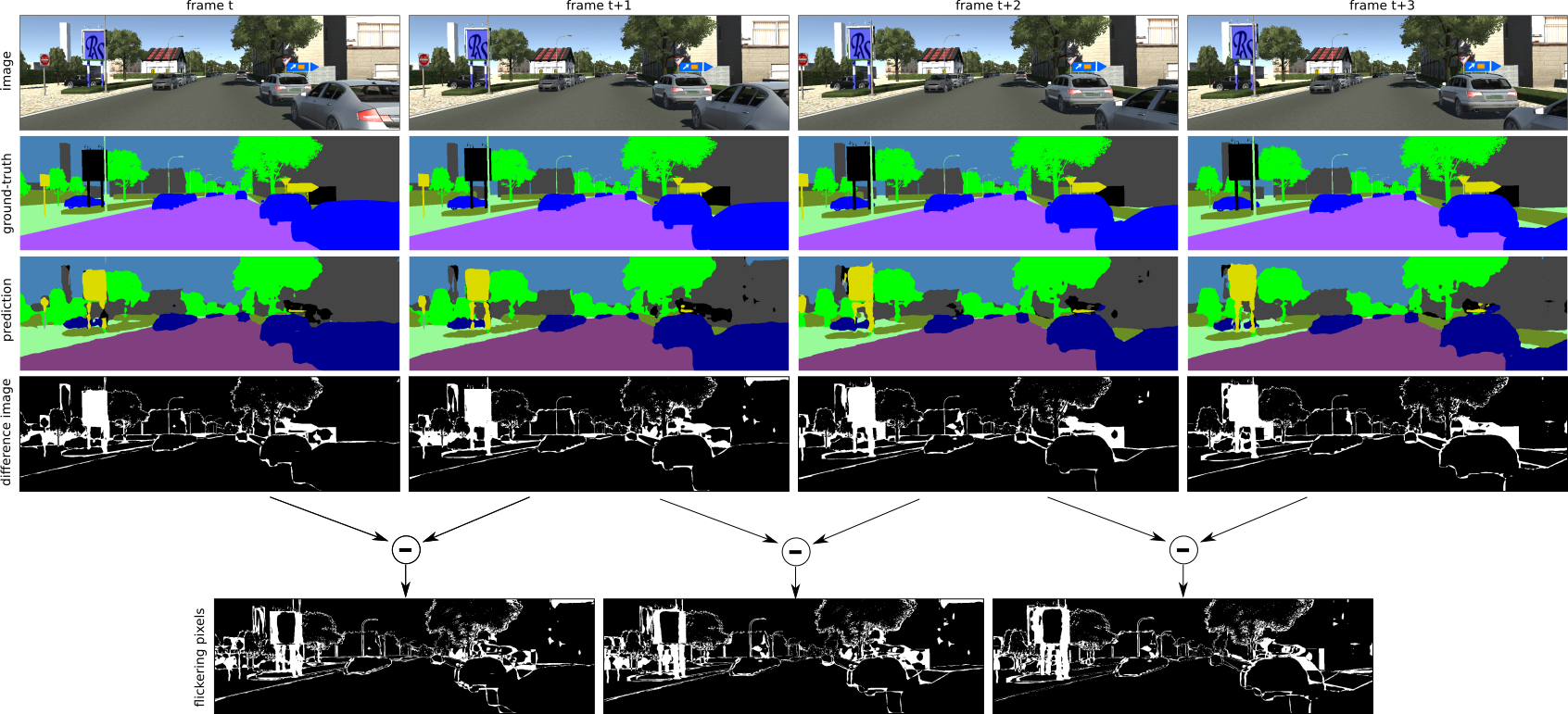}
		\caption{Illustration of mean Flickering Pixels (mFP). First row: images of a video sequence; second row: corresponding ground-truth; third row: yielded segmentation map; fourth row: difference image (difference of segmentation map and ground-truth); fifth row: flickering pixels of two neighboring difference images.}
		\label{fig_illustration_mFP}
		\vspace{-4mm}
	\end{figure*}

\begin{figure*}[t]
		\includegraphics[width=1.0\textwidth]{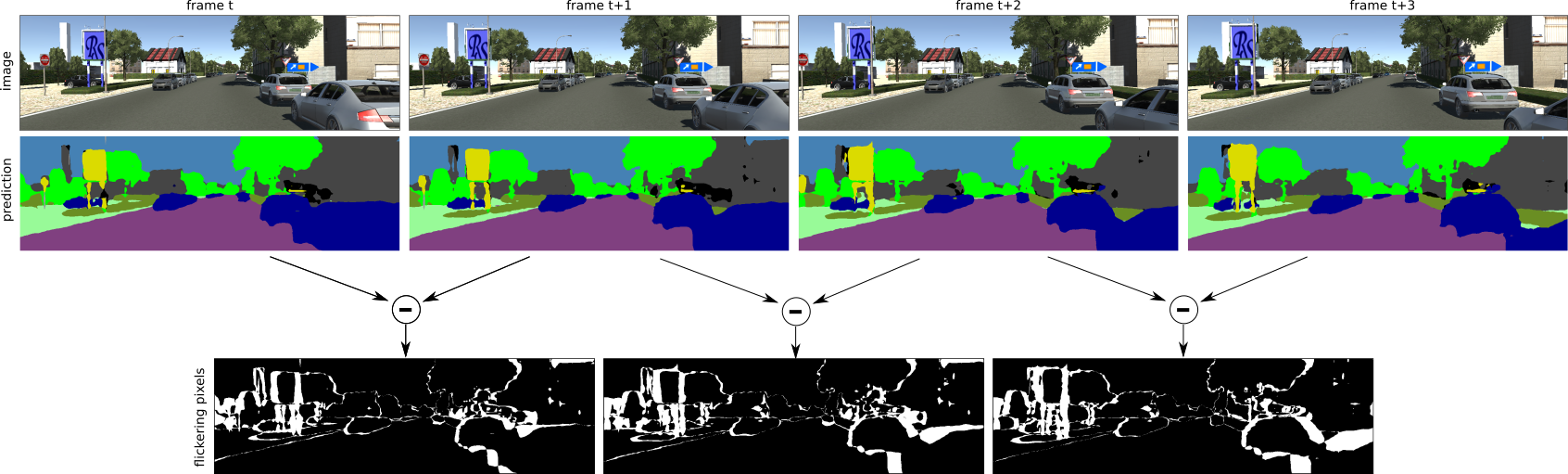}
		\caption{Illustration of mean Flickering Image Pixels (mFIP). First row: images of a video sequence; second row: corresponding segmentation map; third row: flickering pixels of two neighboring frames.}
		\label{fig_illustration_mFIP}
		\vspace{-4mm}
	\end{figure*}

	In video segmentation tasks, several errors only occur in a single frame of the video sequence, and are classified correctly in the following frames. For instance, typical errors are flickering edges and flickering (ghost) objects or object parts. However, these errors can hardly be captured by the conventional evaluation metrics such as pixelwise accuracy and mIoU, since they only consider the results of one time step, and the proportion of these flickering pixels is very small in contrast to the remaining pixels. Hence, an evaluation metric is necessary which measures the flickering pixel errors within a video sequence, and thus, the evaluation metric mean Flickering Pixels (mFP) is introduced, which measures the mismatch of the segmentation result of two neighboring frames. To compensate the motion of moving objects such as walking pedestrians and the ego-motion of the self-driving car between the individual frames, the difference between the result and the corresponding ground-truth is determined first, before the neighboring frames of the video sequence are compared. Furthermore, the mFP is weighted by the amount of image pixels so that the metric is independent of the image size.
	More concretely, the mFP is defined as follows:
	\begin{align}
		mFP = \frac{1}{h \cdot w} \cdot \sum_{t=1}^{T} \; \norm{\mathbf{D}[t] \barwedge \mathbf{D}[t-1]}_1
	\end{align}
	where $\left|\left| \cdot \right|\right|_1$ is the p1-norm, which sums up the absolute values of all matrix elements, $\barwedge$ denotes the elementwise NAND-operator of two matrices, and $h$ and $w$ are the height and the width of the input image, respectively.
	$\mathbf{D}[t]$ is the difference image of the yielded segmentation map $\mathbf{S}[t]$ (\mbox{$\mathbf{S} \in \mathbb{N}^{h \times w}$}, \mbox{$\mathbf{S}_{ij} \in \left\{0, 1, \dots, N-1\right\}$}, and $N$ is the number of classes) and the corresponding ground-truth $\mathbf{G}[t]$ at time step $t$, and is calculated by
	\begin{align}
		\mathbf{D}[t] = \left( \mathbf{S}[t] \barwedge \mathbf{G}[t] \right) \circ \mathbf{S[t]} 
	\end{align}
	The term $\mathbf{S}[t] \barwedge \mathbf{G}[t]$ is multiplied elementwise by the segmentation map $\mathbf{S}[t]$, since $\mathbf{S}[t] \barwedge \mathbf{G}[t]$ is only boolean, and by means of this multiplication, it is predicted which class is different.
	\fig{fig_illustration_mFP} shows the flickering pixels of a short video-sequence according to the mFP-metric.
	The disadvantage of mFP is that the ground-truth for all frames of the video sequence is necessary. However, semantic labeling of each frame is very time-intensive, and so, the ground-truth of only a few frames of the video sequence is often available in common datasets. Therefore, a trimmed-down version of mFP is also proposed, the so called mFIP (mean Flickering Image Pixels), which is independent of the ground-truth. Instead, it measures the mismatch of two neighboring segmentation maps directly. Due to this, moving objects and the ego-motion of the self-driving car cause additional flickering, but it is assumed that this flickering is approximately constant for all evaluated approaches. So, the mFIP is determined by
	\begin{align}
		mFIP = \frac{1}{h \cdot w} \cdot \sum_{t=1}^{T} \; \norm{\mathbf{S}[t] \barwedge \mathbf{S}[t-1]}_1
	\end{align}
	In \fig{fig_illustration_mFIP}, the flickering image pixels of a short video-sequence are shown according to the mFIP-metric.
	All in all, the lower the mFP and mFIP values are, the less flickering points exist in a video sequence.
	Note, that mFP and mFIP do not state anything about the accuracy. For example, if each pixel of a video sequence is classified identically, mFIP is zero, but its performance is very worse.


\section{Evaluation}

	In this section, the proposed convLSTM cells are compared with the standard convLSTM cells by means of the video segmentation task.
	First, the used network architectures are described, and some information about the used datasets and the trainings details are given. Moreover, the proposed convLSTM cells are evaluated by means of different evaluation metrics. Finally, the computational costs and their parameter count are discussed.

\subsection{Network Architectures}

	For the evaluation of the proposed convLSTM cells, three different versions of the LSTM-ICNet introduced in \cite{Pfeuffer_2019_SemanticSegmentationOfVideoSequencesWithConvolutionalLSTMs} are considered, and modified by means of the proposed acceleration techniques to reduce their parameter count and their computational costs. All three LSTM-ICNet versions are extended versions of the ICNet \cite{Zhao_2017_ICNet_forRealTimeSemanticSegmentationOnHighResolutionImages}, where convLSTM layers are added at different positions. In LSTM-ICNet \mbox{version 2}, the convLSTM layer is placed before the softmax layer, and in \mbox{version 5}, a convLSTM layer is added at the end of each resolution branch of the ICNet. LSTM-ICNet \mbox{version 6} is a combination of version 2 and 5, where the convLSTM layers are in front of the softmax layer and at the end of each resolution branch. For more details about the corresponding network architectures see \cite{Pfeuffer_2019_SemanticSegmentationOfVideoSequencesWithConvolutionalLSTMs}.
	The convLSTM cells of all considered versions are replaced by spatial-convLSTM cells, by depth-convLSTM cells and sep-convLSTM cells respectively, and in the following they are compared with the standard versions. 
	Each of the convLSTM cells has a kernel size of $3 \time 3$ and the same number of output channels as its previous layer.

\subsection{Datasets and Training Details}

	In the following, the Cityscapes dataset \cite{cityscape_dataset} and virtual Kitti dataset \cite{virtualKittiDataset} are used for evaluation.
	The Cityscapes dataset consists of $5000$ color images and the corresponding fine-annotated semantic labels. Furthermore, the $19$ previous images are also available for each image of the dataset. Similar to other works \cite{Pfeuffer_2019_RobustSemanticSegmentationInAdverseWeatherConditionsByMeansOfSensorDataFusion, Pfeuffer_2019_SemanticSegmentationOfVideoSequencesWithConvolutionalLSTMs, Zhao_2017_ICNet_forRealTimeSemanticSegmentationOnHighResolutionImages, Zhao_2017_PyramidScenParsingNetwork}, 19 of the 30 available classes are used for evaluation. In the dataset, there are 2975 images for training, and 500 images for validation. The proposed approaches are evaluated on the validation set, since the ground-truth of the test images is not publicly available.
	The virtual Kitti dataset is a photo-realistic synthetic dataset for semantic segmentation containing 50 video sequences in different weather conditions (e.g. rain, fog, sunset), which are 21260 images in total. For each video frame, the corresponding ground-truth is available. Analogously to \cite{Yurdakul_2017_SemanticSegmentationOfRGBDVideosWithRecurrentFullyConvolutionalNeuralNetworks}, the dataset is split into two subsets. The first halves of each video sequence are used for training and the second ones for testing.

	Through all of our experiments, video sequences consisting of four frames are considered, i.e. the temporal image information of the last three frames $t-3$, $t-2$, and $t-1$ are used for determining the segmentation map of frame $t$. 
	The training loss is yielded by means of the cross entropy loss described in \cite{Zhao_2017_ICNet_forRealTimeSemanticSegmentationOnHighResolutionImages}, but the loss is only determined for the last frame of the video sequence similarly to \cite{Pfeuffer_2019_SemanticSegmentationOfVideoSequencesWithConvolutionalLSTMs}, i.e. the result of the previous frames $t-1$, $t-2$, and $t-1$ are not considered for the loss calculation. The training loss is minimized by means of Stochastic Gradient Descent (SGD) with weight decay of $0.0001$ and a momentum equal to $0.9$. Moreover, each model is trained for $30k$, and a poly-learning rate policy is used starting with an initial learning rate of $0.001$. 
	All states of the convLSTM cell are initialized with zero, which corresponds to a complete ignorance of the past, while the remaining parameters are initialized with the same pretrained network. During training, the video sequences are randomly scaled and flipped to avoid overfitting, and the batch size was set to one for Cityscapes and to two for virtual Kitti due to memory reasons.
	Note, that our ICNet implementation differs slightly from its original implementation \cite{Zhao_2017_ICNet_forRealTimeSemanticSegmentationOnHighResolutionImages}, because we skip the model compression afterwards. Instead, we trained with the half feature size. Due to this and due to the small batch size, a mIoU of $60.2\%$ is achieved on Cityscapes (batch size 1), while a mIoU-value of $67.6\%$ (batch size 16) is yielded by the original implementation.

\subsection{Performance Evaluation}

	\begin{table}[t]
		\caption{Evaluation on Cityscapes}
		\centering
		\begin{center}
			\begin{tabular}{|c||c|c|c|}
				\hline
				approach & acc. (\%) & mIoU (\%) & mFIP (\permil)  \Tstrut \Bstrut \\
				\hline \hline \Tstrut
				ICNet \cite{Zhao_2017_ICNet_forRealTimeSemanticSegmentationOnHighResolutionImages} & $92.50$ & $60.07$ & $64.42$ \Mstrut \\
				\hline \Tstrut
				LSTM-ICNet v2  \cite{Pfeuffer_2019_SemanticSegmentationOfVideoSequencesWithConvolutionalLSTMs} & $\mathbf{92.96}$ & $\mathbf{61.93}$ & $62.67$ \Mstrut \\
				spatial-LSTM-ICNet v2  & $92.80$ & $61.70$ & $63.26$ \Mstrut \\
				depth-LSTM-ICNet v2  & $92.56$ & $61.73$ & $61.17$ \Mstrut \\
				sep-LSTM-ICNet v2  & $92.47$ & $60.18$ & $\mathbf{59.77}$ \Mstrut \\
				\hline \Tstrut
				LSTM-ICNet v5  \cite{Pfeuffer_2019_SemanticSegmentationOfVideoSequencesWithConvolutionalLSTMs} & $\mathbf{92.74}$ & $61.53$ & $62.67$ \Mstrut \\
				spatial-LSTM-ICNet v5  & $92.59$ & $\mathbf{61.56}$ & $62.80$ \Mstrut \\
				depth-LSTM-ICNet v5  & $92.52$ & $60.67$ & $61.33$ \Mstrut \\
				sep-LSTM-ICNet v5  & $92.54$ & $60.98$ & $\mathbf{59.74}$ \Mstrut \\
				\hline \Tstrut
				LSTM-ICNet v6  \cite{Pfeuffer_2019_SemanticSegmentationOfVideoSequencesWithConvolutionalLSTMs} & $92.69$ & $\mathbf{61.39}$ & $64.13$ \Mstrut \\
				spatial-LSTM-ICNet v6  & $\mathbf{92.79}$ & $60.90$ & $63.05$ \Mstrut \\
				depth-LSTM-ICNet v6  & $92.30$ & $60.93$ & $61.27$ \Mstrut \\
				sep-LSTM-ICNet v6  & $92.42$ & $60.78$ & $\mathbf{59.72}$ \Mstrut \\
				\hline
			\end{tabular}
		\end{center}
		\label{table_evaluationCityscapes}
	\end{table}

	Now, the different LSTM-ICNet versions described in the previous sections are evaluated by means of pixelwise accuracy (acc.), mIoU, and mFP/mFIP on various datasets. Note, that the number of flickering image pixels (mFIP) is not determined from the validation set of the Cityscapes dataset, since the video-clips are too short. Instead, the mFIP-values are calculated from the demo-videos of this dataset. \tab{table_evaluationCityscapes} and \tab{table_evaluationVirtuellKitti} contain the corresponding results of the Cityscapes dataset and of the virtual Kitti dataset, respectively.
	For both datasets, the standard LSTM-ICNet versions outperform the original ICNet by means of accuracy and mIoU. Furthermore, the amount of flickering pixels is reduced significantly compared to the traditional ICNet. 
	In the previous sections, different acceleration techniques of convLSTM cells were discussed. It turns out that the spatial-LSTM-ICNet versions achieve similar results as the corresponding LSTM-ICNet versions, in some cases they even perform slightly better. In contrast, the depth-LSTM-ICNet and sep-LSTM-ICNet versions perform worse in terms of pixelwise accuracy and mIoU, but they still outperform the origin ICNet in most instances. Nevertheless, there are much less flickering points in the video sequences according to their mFP/mFIP-values than resulting from the origin ICNet.

\subsection{Computational Costs and Memory Consumption}

	\begin{table}[t]
		\caption{Evaluation on Virtual Kitti}
		\centering
		\begin{center}
			\begin{tabular}{|c||c|c|c|}
				\hline
				approach & acc. (\%) & mIoU (\%) & mFP (\permil) \Tstrut \Bstrut \\  
				\hline \hline \Tstrut
				ICNet \cite{Zhao_2017_ICNet_forRealTimeSemanticSegmentationOnHighResolutionImages} & $92.60$ & $58.44$ & $54.42$ \Mstrut \\
				\hline \Tstrut
				LSTM-ICNet v2  \cite{Pfeuffer_2019_SemanticSegmentationOfVideoSequencesWithConvolutionalLSTMs} & $\mathbf{93.01}$ & $59.71$ & $51.55$ \Mstrut \\
				spatial-LSTM-ICNet v2  & $92.89$ & $59.55$ & $\mathbf{51.33}$ \Mstrut \\
				depth-LSTM-ICNet v2  & $92.86$ & $\mathbf{59.77}$  & $51.70$ \Mstrut \\
				sep-LSTM-ICNet v2  & $92.68$ & $58.72$ & $52.16$ \Mstrut \\
				\hline \Tstrut
				LSTM-ICNet v5  \cite{Pfeuffer_2019_SemanticSegmentationOfVideoSequencesWithConvolutionalLSTMs} & $92.96$ & $60.19$ & $51.66$ \Mstrut \\
				spatial-LSTM-ICNet v5  & $\mathbf{93.02}$ & $\mathbf{60.37}$ & $\mathbf{51.27}$ \Mstrut \\
				depth-LSTM-ICNet v5  & $92.56$ & $59.09$ & $52.56$ \Mstrut \\
				sep-LSTM-ICNet v5  & $92.57$ & $58.67$ & $52.22$ \Mstrut \\
				\hline \Tstrut
				LSTM-ICNet v6  \cite{Pfeuffer_2019_SemanticSegmentationOfVideoSequencesWithConvolutionalLSTMs} & $\mathbf{93.07}$ & $60.50$ & $\mathbf{50.90}$ \Mstrut \\
				spatial-LSTM-ICNet v6  & $92.99$ & $\mathbf{60.54}$ & $51.10$ \Mstrut \\
				depth-LSTM-ICNet v6  & $92.53$ & $59.03$  &  $53.52$ \Mstrut \\
				sep-LSTM-ICNet v6  & $92.64$ & $58.85$ & $52.99$ \Mstrut \\
				\hline
			\end{tabular}
		\end{center}
		\label{table_evaluationVirtuellKitti}
	\end{table}
	
	\begin{table*}[t]
		\caption{Comparison of Memory, FLOPs, and inference time}
		\centering
		\begin{center}
			\begin{tabular}{|c||c|c||c|c||c|c||c|c|}
				\hline \Tstrut
				approach & \multicolumn{2}{|c||}{model parameters} & \multicolumn{2}{|c||}{GFLOPs} & \multicolumn{2}{|c||}{inference time (CPU)} & \multicolumn{2}{|c|}{inference time (GPU)} \Mstrut \\
				& amount & percentage & amount & percentage & time & percentage  & time & percentage  \Tstrut \Bstrut \\
				\hline \hline \Tstrut
				ICNet \cite{Zhao_2017_ICNet_forRealTimeSemanticSegmentationOnHighResolutionImages} & $6707k$ & $100.00\%$ & $58.38$ & $100.00\%$ & $1388ms$ & $100.00\%$ & $48.19 ms$ & $100.00\%$ \Mstrut \\
				\hline \Tstrut
				LSTM-ICNet v2  \cite{Pfeuffer_2019_SemanticSegmentationOfVideoSequencesWithConvolutionalLSTMs} & $7887k$ & $100.00\%$ & $135.74$ & $100.00\%$ & $2503ms$ & $100.00\%$ & $65.01ms$ & $100.00\%$ \Mstrut \\
				spatial-LSTM-ICNet v2  &  $7494k$ & $95.02\%$ & $109.97$ & $81.01\%$ & $2265ms$ & $90.47\%$ & $68.15ms$ & $104.83\%$ \Mstrut \\
				depth-LSTM-ICNet v2  & $\mathbf{6717k}$ & $\mathbf{85.17\%}$ & $\mathbf{59.03}$ & $\mathbf{43.49\%}$ & $\mathbf{1616ms}$ & $\mathbf{64.56\%}$ & $\mathbf{60.62ms}$ & $\mathbf{93.24\%}$ \Mstrut \\
				sep-LSTM-ICNet v2  & $6848k$ & $86.83\%$ & $67.62$ & $49.82\%$ & $1734ms$ & $69.26\%$ & $62.18ms$ & $95.65\%$ \Mstrut \\
				\hline \Tstrut
				LSTM-ICNet v5  \cite{Pfeuffer_2019_SemanticSegmentationOfVideoSequencesWithConvolutionalLSTMs} & $10247k$ & $100.00\%$ & $173.98$ & $100.00\%$ & $2967ms$ & $100.00\%$ & $69.64ms$ & $100.00\%$ \Mstrut \\
				spatial-LSTM-ICNet v5  & $9068k$ & $88.49\%$ & $135.4$ & $77.87\%$ & $2565ms$ & $86.45\%$ & $74.82ms$ & $107.43\%$ \Mstrut \\
				depth-LSTM-ICNet v5  & $\mathbf{6736k}$ & $\mathbf{65.74\%}$ & $\mathbf{59.36}$ & $\mathbf{34.12\%}$ & $\mathbf{1664ms}$ & $\mathbf{56.07\%}$ & $\mathbf{62.65ms}$ & $\mathbf{89.96\%}$ \Mstrut \\
				sep-LSTM-ICNet v5  & $7129k$ & $69.57\%$ & $72.20$ & $41.50\%$ & $1828ms$ & $61.61\%$ & $65.08ms$ & $93.46\%$ \Mstrut \\
				\hline \Tstrut
				LSTM-ICNet v6  \cite{Pfeuffer_2019_SemanticSegmentationOfVideoSequencesWithConvolutionalLSTMs} & $11428k$ & $100.00\%$ & $251.34$ & $100.00\%$ & $4125ms$ & $100.00\%$ & $80.16ms$ & $100.00\%$ \Mstrut \\
				spatial-LSTM-ICNet v6 & $9855k$ & $86.24\%$ & $187.07$ & $74.43\%$ & $3403ms$ & $82.49\%$ & $88.08ms$ & $109.87\%$ \Mstrut \\
				depth-LSTM-ICNet v6  & $\mathbf{6746k}$ & $\mathbf{59.03\%}$ & $\mathbf{60.02}$ & $\mathbf{23.88\%}$ & $\mathbf{1879ms}$ & $\mathbf{45.55\%}$ & $\mathbf{68.60ms}$ & $\mathbf{85.57\%}$ \Mstrut \\
				sep-LSTM-ICNet v6  & $7270k$ & $63.62\%$ & $81.44$ & $32.40\%$ & $2127ms$ & $51.55\%$ & $72.68ms$ & $90.66\%$ \Mstrut \\
				\hline
			\end{tabular}
		\end{center}
		\vspace{-5mm}
		\label{table_memoryComputationTime}
	\end{table*}

	In this section, the computational costs and the memory consumption of the proposed approaches are discussed. Each approach was implemented in Tensorflow \cite{tensorflow}, and executed on a single GPU (Nvidia TitanX) and on a CPU (Intel(R) Core(TM) i5-6300U CPU @ 2.40GHz). \tab{table_memoryComputationTime} contains the corresponding results using an input image of size $1024 \times 2048$. The experiments show that the number of FLOPs are reduced by means of the proposed spatial-convLSTMs, depth-convLSTMs, and spatial-convLSTMs. As expected, the depth-convLSTMs versions require the least number of FLOPs, since depthwise convolutions are more computational efficient than spatial and depthwise separable convolutions. Furthermore, the percentage reduction of the FLOPs is greatest for the modified LSTM-ICNet version 6, because these versions contain most convLSTM cells, and hence, they can be sped up most by means of the proposed separable convLSTM cells. 
	Due to the reduced number of FLOPs, the modified LSTM-ICNet versions take much less computation time on the CPU than their origin versions, as the results in \tab{table_memoryComputationTime} show. For example, the inference time of LSTM-ICNet version 6 is approximately halved if the convLSTM cells are replaced by depth-convLSTM cells or sep-convLSTM cells. The inference time is also reduced on the GPU, but not as much as on the CPU, and the spatial convLSTM versions take even longer than the original LSTM-ICNet versions. The reason is that the standard convolution operations are highly optimized in deep-learning frameworks, especially for $3\times3$ kernels so that the overhead of doing two convolutions on the GPU outweighs its speedup for small kernel sizes. For greater kernel sizes, the spatial convLSTM cells are again faster than the conventional convLSTM cells. The spatially separable convolutions can be surely implemented more efficiently by means of computational tricks such as \cite{Lavin_FastAlgorithmsForConvolutionalNeuralNetworks} to reduce their execution time, however, this is out of the scope of this work. 
	Additionally, the required memory decreases for spatial-LSTM-ICNet, depth-LSTM-ICNet and sep-LSTM-ICNet, since their parameter count is reduced by up to $41\%$. The parameter saving is greatest for the depth-LSTM-ICNet versions, while it is lowest for the spatial-LSTM-ICNet versions.


\section{Conclusion}

	In this paper, three different approaches to speed up standard convLSTM cells were investigated by the video segmentation task. It turned out, that the spatial convLSTM cells achieve similar performance on well-known datasets than the standard convLSTM approaches, and are more computational efficient and require less model parameters at the same time.
	The number of FLOPs and the parameter count can be increased further by using depthwise or depthwise separable convLSTM cells instead, but at the expense of performance. 
	Furthermore, the evaluation metric mean Flickering Pixels (mFP) was introduced, which measures the number of flickering pixels within a video sequence. Experiments with the proposed separable convLSTM cells show, that the amount of flickering pixels can be reduced significantly, if temporal image information are also considered for segmenting video sequences by means of recurrent neural networks.

\section{Acknowledgment}
	
	The research leading to these results has received funding from the European Union under the H2020 ECSEL Programme as part of the DENSE project, contract number 692449.

\bibliographystyle{plain}
\bibliography{/home/andreas/Documents/Literatur/Jabref-Datebase/Literatur_Promotion}

\end{document}